\pdfoutput=1
\documentclass[]{smule}
\usepackage[round,authoryear]{natbib}
\usepackage{hyperref}
\usepackage[colorinlistoftodos]{todonotes}
\usepackage{colortbl}
\usepackage{nicematrix}
\usepackage{amssymb}
\usepackage{amsmath}
\usepackage{booktabs}
\usepackage{adjustbox}
\usepackage{soul}
\usepackage[inline]{enumitem}
\usepackage{booktabs}
\usepackage{color}
\usepackage{xcolor}
\usepackage{bbding}
\usepackage{listings}
\usepackage{tikz}
\usepackage{multirow}
\usepackage{pgfplots}
\usepackage{graphicx}
\usepackage{array}
\usepackage{rotating}
\usepackage{multicol}
\usepackage{xspace}
\usepackage{tablefootnote}
\usepackage{libertinus}

\makeatletter
\AtBeginDocument{

}
\makeatother

\title{Learning Interpretable Features in Audio Latent Spaces via Sparse Autoencoders}

\author[3\dagger\ddagger]{Nathan Paek}
\author[1\ddagger]{Yongyi Zang}
\author[2\dagger\ddagger]{Qihui Yang}
\author[1]{Randal Leistikow}

\affiliation[1]{Smule Labs}
\affiliation[2]{University of California, San Diego}
\affiliation[3]{Stanford University}

\contribution[\dagger]{Work done during internship at Smule}
\contribution[\ddagger]{These authors contributed equally.}

\abstract{
While sparse autoencoders (SAEs) successfully extract interpretable features from language models, applying them to audio generation faces unique challenges: audio's dense nature requires compression that obscures semantic meaning, and automatic feature characterization remains limited. We propose a framework for interpreting audio generative models by mapping their latent representations to human-interpretable acoustic concepts. We train SAEs on audio autoencoder latents, then learn linear mappings from SAE features to discretized acoustic properties (pitch, amplitude, and timbre). This enables both controllable manipulation and analysis of the AI music generation process, revealing how acoustic properties emerge during synthesis. We validate our approach on continuous (DiffRhythm-VAE) and discrete (EnCodec, WavTokenizer) audio latent spaces, and analyze DiffRhythm, a state-of-the-art text-to-music model, to demonstrate how pitch, timbre, and loudness evolve throughout generation. While our work is only done on audio modality, our framework can be extended to interpretable analysis of visual latent space generation models.
}

\begin{document}

\maketitle

\section{Introduction}

As powerful neural networks become more integrated into society, their lack of interpretability raises a significant concern~\citep{hendrycks2023overview}. To address this challenge, sparse autoencoders (SAEs) have emerged as a key tool in mechanistic interpretability research~\citep{olah2020zoom, cammarata2021curve, nelson2021mathematical}. They are motivated by the polysemantic hypothesis~\citep{olah2020zoom, elhage2022toy, marshall2024understanding}: that neurons encode more features than dimensions by superposing multiple concepts. SAEs work by finding sparse directions in activation space to isolate these underlying, disentangled features. 
This approach has proven effective in large language models (LLMs), where SAEs can extract highly monosemantic features that are automatically characterized by using the model itself to summarize the results of token-level perturbations~\citep{cunningham2023sparseautoencodershighlyinterpretable}.


However, extending this approach to audio generative networks presents fundamental challenges. Unlike text, audio is inherently dense~\citep{wu2024towards}, and thus typically requires learned compression through autoencoders before tokenization~\citep{liu2023audioldm}. This compression step, whether producing continuous or discrete latent codes, obscures the semantic meaning of individual ``tokens,'' making perturbation-based analysis less interpretable~\citep{wu2024towards, ye2025codec}. Moreover, while language models excel at summarizing textual patterns, current audio understanding models are not yet capable of providing an equally robust automatic characterization of SAE feature behaviors~\citep{su2025audio, yang2025towards}. These limitations necessitate new approaches for interpretable feature discovery in audio generative systems.

In this work, we propose a novel framework for understanding audio generative models by analyzing their latent space representations through human-interpretable acoustic concepts. Our approach proceeds in three stages. First, we train SAEs on the latent representations of audio autoencoders to extract sparse features. Second, we learn linear mappings from these SAE features to human-interpretable acoustic concepts: pitch, amplitude, and timbre (represented here by spectral centroid as a simplified proxy~\citep{schubert2004spectral, schubert2006does}). To enable discrete analysis, we quantize each acoustic property into interpretable ``units'': pitch is discretized according to the Western tonal system (e.g., C4, C\#4), while amplitude and spectral centroid are binned with equal spacing within their physical ranges. 
The effectiveness of linear mappings suggests that SAE features already encode acoustic properties in a near-linear fashion, validating the hypothesis that these learned representations align with human-interpretable concepts. Finally, by decomposing the audio synthesis process into an interpretable feature hierarchy, our framework traces how specific acoustic properties emerge. We empirically validate this approach on DiffRhythm, a state-of-the-art text-to-music model. 
Although our experiments focus on audio, we believe this framework is generalizable to other generative models that operate within learned latent spaces, including those for image and video.



\begin{figure*}
    \centering
    \includegraphics[width=\linewidth]{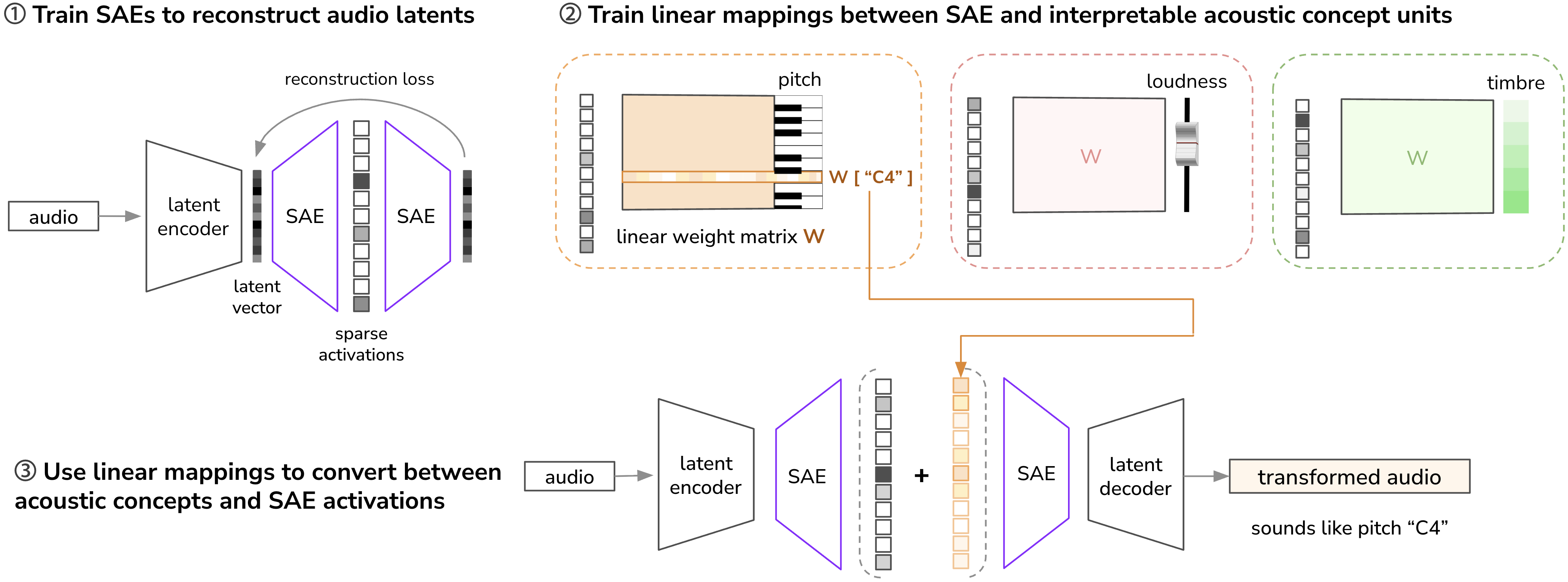}
    \caption{Framework for interpreting and controlling audio generative models through sparse features learned on their generation space. Sparse autoencoders extract interpretable features from audio latents, which are then linearly mapped to acoustic concepts. Control vectors extracted from these linear mappings can then be used to transform audio.}
    \label{fig:figure1}
\end{figure*}

\section{Methodology}
\subsection{Sparse Autoencoder Training}
We train SAEs on latent representations from three pretrained audio encoders: the continuous VAE space of Stable Audio Open and DiffRhythm \citep{sao, ning2025diffrhythmblazinglyfastembarrassingly}, and the discrete latent spaces of EnCodec \citep{défossez2022highfidelityneuralaudio} and WavTokenizer \citep{ji2025wavtokenizerefficientacousticdiscrete}. To address the unique requirements of audio latents, we modify the standard SAE architecture by adding an RMS normalization layer after the ReLU activation. This modification maintains consistent activation magnitudes and, as we empirically found, prevents out-of-distribution artifacts during feature manipulation. Following standard practice \citep{cunningham2023sparseautoencodershighlyinterpretable}, we optimize the SAEs using a composite loss function:
\begin{equation}
\mathcal{L} = \|\mathbf{x} - \hat{\mathbf{x}}\|_2^2 + \lambda \|\mathbf{h}\|_1
\end{equation}
where the first term ensures reconstruction fidelity and the $L_1$ penalty promotes sparsity in the hidden activations $\mathbf{h}$. We conduct systematic grid searches over hidden dimensionalities (ranging from $4\times$ to $256\times$ the input dimension) and sparsity coefficients $\lambda$ (ranging from 0.005 to 0.15) to identify optimal configurations for each latent space.

\subsection{Linear Mapping to Acoustic Concepts}
To connect SAE features to interpretable acoustic properties, we train linear probes that predict discretized audio attributes from sparse activations. Given a latent vector $\mathbf{x} \in \mathbb{R}^{d}$, our SAE produces sparse features:
\begin{equation}
\mathbf{h} = \text{ReLU}(W_{\text{enc}} \mathbf{x} + \mathbf{b}_{\text{enc}}), \quad
\mathbf{f} = \text{RMSNorm}(\mathbf{h})
\end{equation}
where $\mathbf{f} \in \mathbb{R}^{m}$ are the normalized features used for both reconstruction and interpretation.

\begin{figure*}
    \centering
    \includegraphics[width=\linewidth]{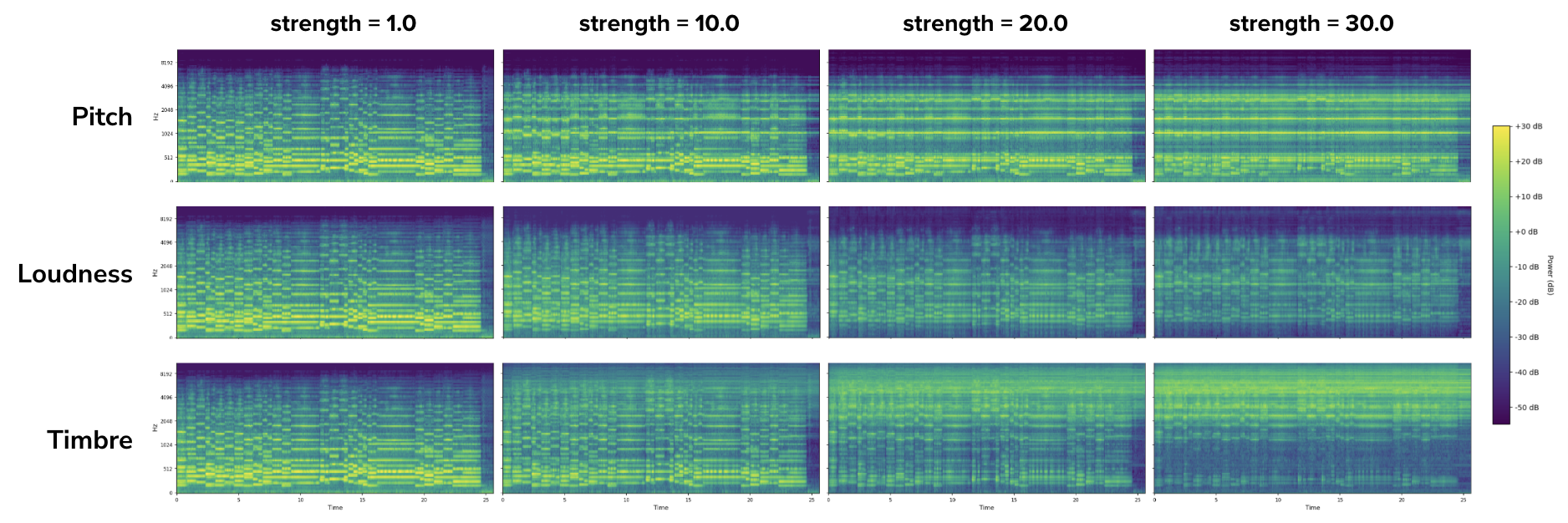}
    \caption{Controlled audio manipulation via control vectors. When $\alpha$ increases, isolated changes in pitch (imminent C5), amplitude (decreasing loudness), and timbre (brightening via high-frequency emphasis) can be observed. \textcolor{red}{Corresponding audio samples can be found here: \url{https://anonymous.4open.science/r/audio_samples-A301/}}}
    \label{fig:figure2}
\end{figure*}

For each acoustic attribute $a \in \{\text{pitch}, \text{amplitude}, \text{timbre}\}$, we first extract continuous measurements from the audio: pitch via CREPE~\citep{kim2018crepe}, amplitude via windowed RMS energy using librosa \citep{mcfee2025librosa}, and timbre via windowed spectral centroid using librosa. We then discretize these continuous curves into $K_a$ classes (pitch using logarithmic bins aligned with MIDI note numbers, and amplitude/timbre using linear bins) and train a linear classifier:
\begin{equation}
p^{(a)} = \text{softmax}(W^{(a)} \mathbf{f} + \mathbf{b}^{(a)})
\end{equation}
where $W^{(a)} \in \mathbb{R}^{K_a \times m}$ maps SAE features to class logits. The linearity provides bidirectional interpretability, as the contribution of SAE feature $j$ to acoustic class $k$ is simply $c_{j \to k}^{(a)} = W^{(a)}_{kj} \cdot f_j$, where the weights $W^{(a)}_{kj}$ reveal both which features encode specific acoustic properties and how acoustic concepts decompose into SAE features. For targeted intervention, we leverage this linearity directly by adding the scaled probe weight vector $\alpha \cdot \mathbf{w}k^{(a)}$ (\textit{"control vectors"}) to the SAE features to shift the audio toward acoustic class $k$. After re-normalizing to maintain valid activation magnitudes, we decode through both SAE and audio decoders to generate the modified audio (shown in Figure \ref{fig:figure2}). 

\section{Experiments}

\subsection{Acoustic Concept Mapping Discovery}
\textbf{Dataset.} We use a composite dataset of $\sim$31 hours of audio sampled from several sources: CocoChorales \citep{wu2022chamber}---11.2 hours of four-part Bach chorales, DAMP-VSEP \citep{smule2019dampvsep}---11.7 hours of pop/rock singing, the Extended Groove MIDI Dataset \citep{callender2020improving}---7.8 hours of drums, GuitarSet \citep{xi2018guitarset}---24 minutes of solo guitar, and MAESTRO \citep{hawthorne2018enabling}---33 minutes containing classical piano.

\textbf{Training SAEs on audio latent spaces.} We conduct  grid searches over SAE hidden dimensions $\{2048, 4096, 8192, 12288, 16384\}$ and sparsity coefficients $\lambda \in \{0.005, 0.01, 0.05, 0.1, 0.15\}$ for each audio encoder. The resulting SAEs exhibit distinct characteristics across latent spaces. DiffRhythm achieves sparsity ratios ranging from 0.65 to 0.98. WavTokenizer produces the sparsest representations (0.993--0.999), suggesting its discrete tokens already encode highly disentangled features. EnCodec demonstrates the widest sparsity range (0.55--0.95). Across all models, larger hidden dimensions consistently improve reconstruction quality.

\begin{figure*}
    \centering
    \includegraphics[width=\linewidth]{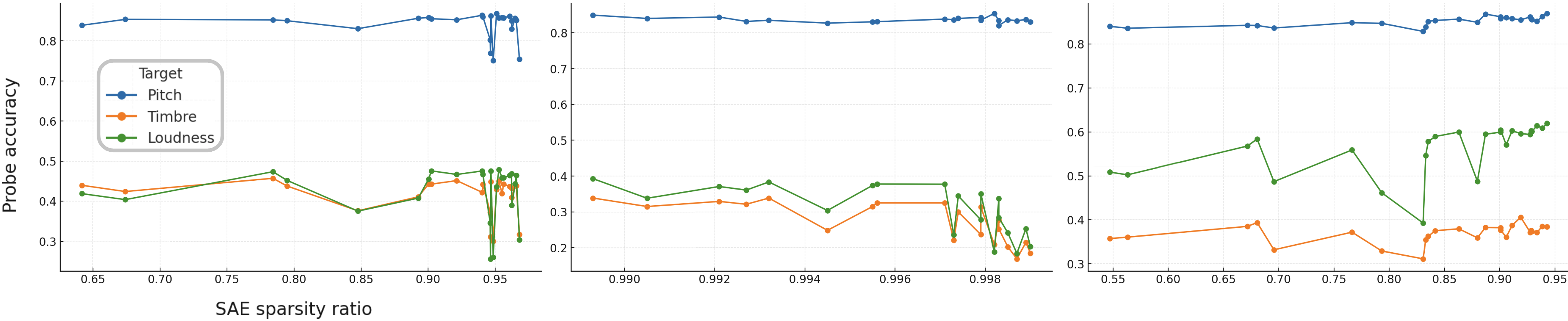}
    \caption{Linear probe accuracy for acoustic property classification across different sparsity levels. Left: Stable Audio Open/DiffRhythm VAE, Middle: WavTokenizer, Right: EnCodec.}
    \label{fig:figure3}
\end{figure*}

\textbf{Training linear probes from SAE features to acoustic concepts.} We train linear probes to predict pitch (with 66 bins spanning the pitch range present in our dataset), loudness (20 bins), and timbre (20 bins) from SAE features. Plotting the probe classification accuracy on a test set vs. the sparsity of its SAE in Figure~\ref{fig:figure3} shows a hierarchy of linear decodability across acoustic properties. Pitch proves most linearly separable (0.75--0.87 accuracy) and remains stable across all sparsity levels, suggesting fundamental frequency encoding. EnCodec excels at loudness (0.56--0.63) compared to DiffRhythm and WavTokenizer (0.17--0.49). Timbre remains challenging across all models (0.17--0.46).

\textbf{Applying targeted interventions to audio samples.} We test controllability on diverse audio sources (singing voice, drums, four-part harmony). Figure~\ref{fig:figure2} shows a chordal audio sample from the CocoChorales dataset encoded with EnCodec and our highest-sparsity SAE (hidden\_dim=16384, $\lambda=0.1$). We apply control vectors targeting pitch (MIDI C5), timbre (spectral centroid class 17), and loudness (class 2) with strengths $\alpha \in \{1, 10, 20, 30\}$. As $\alpha$ increases, edits are isolated in the targeted attribute, while non-targeted properties remain largely preserved.

\subsection{Generation Process Visualization}
\begin{figure*}
    \centering    
    \includegraphics[width=0.8\linewidth]{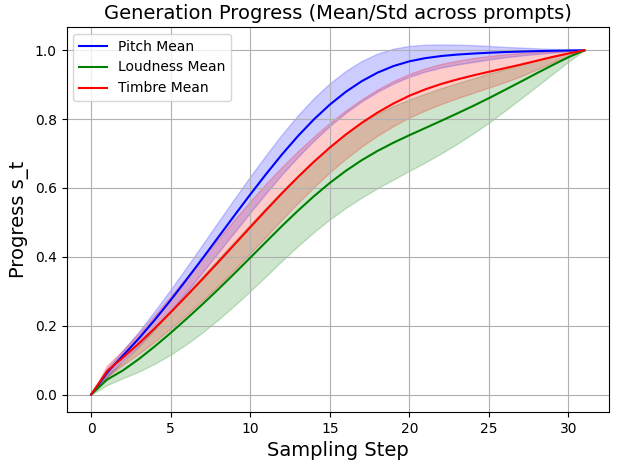}
    \caption{Probes Variation in Generation Progress }
    \label{fig:generation-progress}
\end{figure*}

We demonstrate how our learned mappings can help us understand the audio generation process by analyzing DiffRhythm~\citep{ning2025diffrhythmblazinglyfastembarrassingly}, a rectified flow model designed for full-length song synthesis. In this analysis, the model was configured to generate a 95-second audio segment, encompassing a verse and a chorus, over 32 inference steps. At each generation step $t \in \{0, ..., 31\}$, we extract the latent $\mathbf{X}_t\in \mathbb{R}^{C \times F}$, and decompose it through our SAE and linear probes to obtain acoustic concept activations $\mathbf{P}^{(a)}_t\in \mathbb{R}^{F \times K_a}$.  After applying a mean pooling over frames ($F$), we obtain distributions $p^{(a)}_t$ for each attribute $a$. To quantify the evolution of acoustic properties, we track how these distributions interpolate from noise to the final audio. Specifically, for each attribute $a$ and step $t$, we compute the per-class normalized $L^1$ distance:
\begin{equation}
s^{(a)}_t = \frac{1}{K_a} \sum_{k=1}^{K_a} \frac{|p^{(a)}_{t,k} - p^{(a)}_{0,k}|}{|p^{(a)}_{T,k} - p^{(a)}_{0,k}|}
\end{equation}
where $s^{(a)}_t \in [0, 1]$ measures the progression from initial noise ($t=0$) toward the final acoustic structure ($t=T=31$). This reveals when different acoustic properties emerge during generation. We sample 500 prompts from MusicCaps~\citep{agostinelli2023musiclm}, then plot the mean and standard deviation of the generation progress in Figure~\ref{fig:generation-progress}, which indicates a clear hierarchy. Pitch converges first (around step 21), followed by timbre, while loudness converges last and remains unresolved by the final step. This coarse-to-fine progression suggests the model establishes fundamental frequency before refining textural and dynamic details.

\section{Conclusion}

We present a framework for interpreting audio generative models by mapping their latent representations to human-interpretable acoustic properties through sparse autoencoders and linear probes. Our experiments demonstrate that SAE features naturally align with acoustic properties, enabling both controllable manipulation and better understanding of music generative models.

In future work, we plan to apply our method to other generative architectures, such as RAVE \citep{caillon2021ravevariationalautoencoderfast}, ACE-Step \citep{gong2025acestepstepmusicgeneration}, and AudioLDM \citep{liu2023audioldm}. Beyond the three acoustic properties explored here, we will train probes for richer audio features such as rhythm, harmony, and instrument identity. Finally, we aim to use these interpretable features to directly guide generation behavior during inference, potentially enabling fine-grained control over specific attributes while maintaining generation quality. 

\clearpage
\newpage
\bibliographystyle{plainnat}
\bibliography{paper}

\end{document}